\documentclass[10pt,twocolumn,letterpaper]{article}

\usepackage[pagenumbers]{cvpr} %

\usepackage{graphicx}
\usepackage{amsmath}
\usepackage{amssymb}
\usepackage{booktabs}

\usepackage[pagebackref,breaklinks,colorlinks]{hyperref}

\usepackage[capitalize]{cleveref}
\crefname{section}{Sec.}{Secs.}
\Crefname{section}{Section}{Sections}
\Crefname{table}{Table}{Tables}
\crefname{table}{Tab.}{Tabs.}

\usepackage{xcolor}

\usepackage{soul}
\usepackage{scrextend}

\newcommand{\ourmethod}{{T3AR }}

\def\cvprPaperID{10053} %
\def\confName{CVPR}
\def\confYear{2022}

\def\ourname{${\rm T^3AR}$ }
\def\ournamenospace{${\rm T^3AR}$}

\begin{document}

\title{Train/Test-Time Adaptation with Retrieval}

\author{
Luca Zancato
\quad
Alessandro Achille
\quad
Tian Yu Liu\thanks{Work done during an internship at AWS AI Labs.}
\quad
Matthew Trager\\[.15cm]
Pramuditha Perera
\quad
Stefano Soatto\vspace{0.2em} \\[.2cm]
AWS AI Labs \vspace{0.2em}\\[.05cm]
{\tt zancato@amazon.it}\\
{\tt\{aachille,tytianyu,mttrager,pramudi,soattos\}@amazon.com}\\
}
\maketitle

\begin{abstract}
    We introduce Train/Test-Time Adaptation with Retrieval~(\ournamenospace), a method to adapt models both at train and test time by means of a retrieval module and a searchable pool of external samples. Before inference, \ourname adapts a given model to the downstream task using refined pseudo-labels and a self-supervised contrastive objective function whose noise distribution leverages retrieved real samples to improve feature adaptation on the target data manifold. 
    The retrieval of real images is key to \ourname since it does not rely solely on synthetic data augmentations to compensate for the lack of adaptation data, as typically done by other adaptation algorithms.
    Furthermore, thanks to the retrieval module, our method gives the user or service provider the possibility to improve model adaptation on the downstream task by incorporating further relevant data or to fully remove samples that may no longer be  available due to changes in user preference after deployment.
    First, we show that \ourname can be used at training time to improve downstream fine-grained classification over standard fine-tuning baselines, and the fewer the adaptation data the higher the relative improvement (up to 13\%). 
    Second, we apply \ourname for test-time adaptation and show that exploiting a pool of external images at test-time leads to more robust representations over existing methods on DomainNet-126 and VISDA-C, especially when few adaptation data are available (up to 8\%). 
\end{abstract}

\begin{figure}
    \centering
    \includegraphics[width=\linewidth]{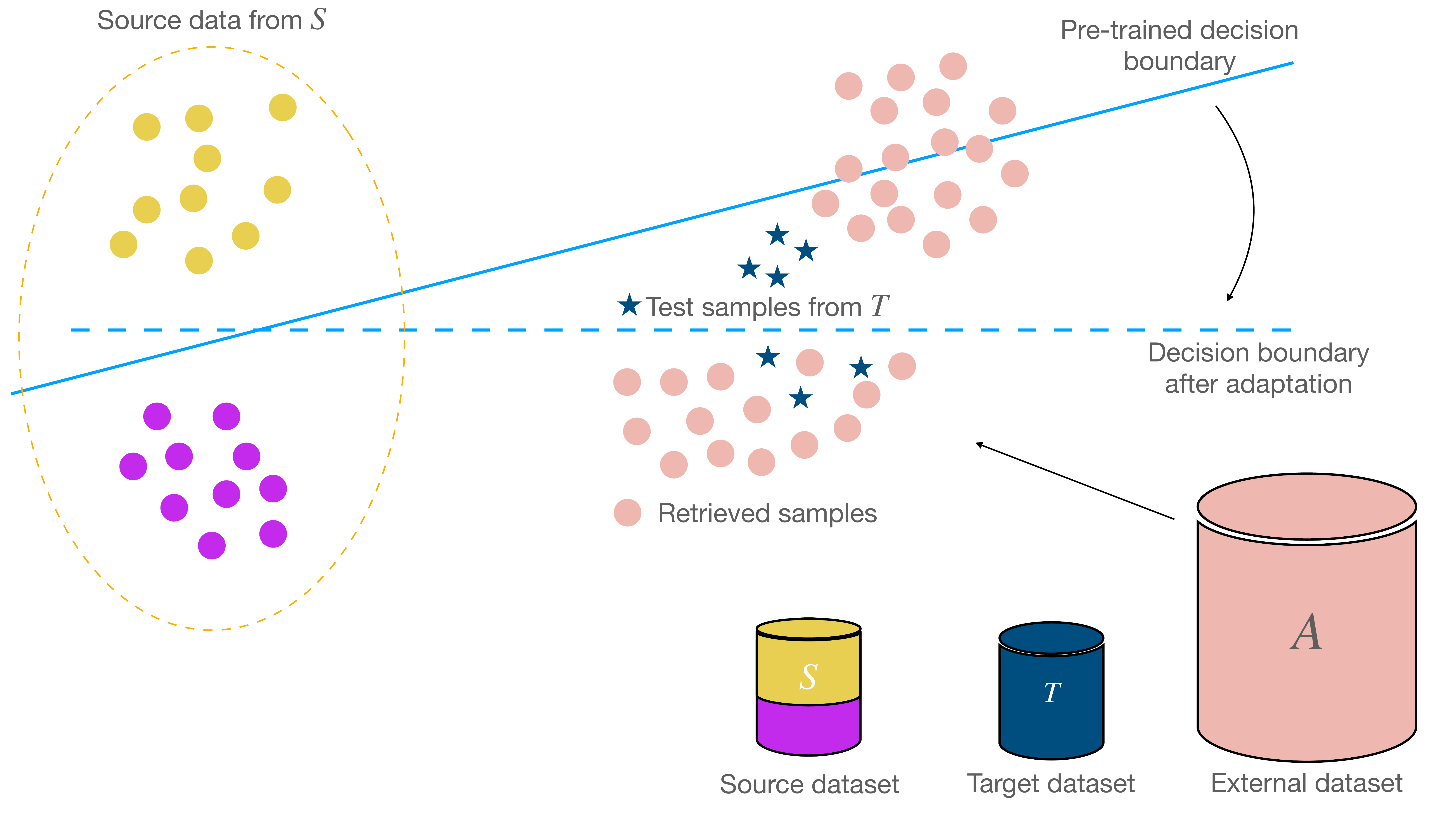}
    \caption{
    \textbf{Adaptation with retrieval from an external data pool.} Illustration of how \ourname exploits target data $T$ and the external data pool $A$ to adapt the decision boundary after pre-training on the source datasets $S$. For new test queries from the target dataset, \ourname approximates the local data manifold around $T$ by retrieving similar unlabelled examples from $A$. Then, it updates the decision boundary with a contrastive self-supervised objective. 
    }
    \label{fig:fpl}
\end{figure}

\section{Introduction}\label{sec:intro}

While Deep Learning models are evolving rapidly, machine learning systems used in production are updated rarely, as each deployment requires the provider to engage in a complex process of scaling, securitization, certification of new model and dataset cards, bias evaluation, and regression tests.  It is now common for users to adapt trained models to their specific use cases, or to the changed context as time goes by \cite{continual_learning_survey, online_continual_learning, prompt_continual_learning}. Such adaptation can be performed by fine-tuning on a specific dataset $S$ owned by the user \cite{model_zoo, LQF}. However, on an even finer time-scale, users may want to adapt their models based on data they observe at test time, bypassing the time-consuming annotation process \cite{adaContrast, tent, test_time_training, SHOT}. Test-Time Adaptation (TTA) refers to the problem of adapting a source model to a target task $T$ represented by test data, for which no ground-truth labels are given.

This trend is exacerbated by the advent of Foundation Models \cite{OFA, unifiedio, pixel2seq, flamingo}, at least in the visual domain where tasks can be antagonistic and models are sensitive to even subtle changes in the data distribution. At the same time, both users and providers typically have access to ever-growing pools of auxiliary data, albeit often heterogeneous (pertaining to concepts other than the one of interest at test-time), and without annotations. Yet it seems plausible that, somewhere within these large pools of data, there may be information useful for the task at hand. 

In this paper, we tackle the problem of performing test-time adaptation by retrieving information from a large, unlabeled, heterogeneous, and evolving dataset. 
The same procedure could also be followed by the provider, if they have access to auxiliary internal data and wish to adapt the production model based on trends observed in test data. 
We refer to our method as {\em Train/Test-Time Adaptation with Retrieval}, or \ournamenospace.

 \ournamenospace, if solved, would enable a number of real-world tasks that have thus far frustrated practitioners. For instance, it would allow a user to select, among a vast data lake $A$, which samples to use for a training, based on labeled and unlabeled samples~\cite{yan2020neural}. It would also enable nimble inference, by adapting a modest-size model to specific tasks, rather than relying on an unwieldy model to master all trades. Finally, it would enable {\em reversible adaptation:} While in the case of language models tasks are generally synergistic \cite{tanl}, in vision tasks can be antagonistic.\footnote{{\em E.g.}, localization requires marginalizing identity, whereas recognition requires marginalizing location, making the features that are informative for one detrimental to the other \cite{flamingo, unifiedio}.} Therefore, a model adapted to one task may behave poorly on another, and a model that encompasses both would require significantly higher capacity \cite{OFA, unifiedio, pixel2seq, flamingo}, to the detriment of inference efficiency. In \ournamenospace, changing the target data $T$ changes the subset of the data pool $A$ that is retrieved, with no impact on other models, instantiating smaller independent models for antagonistic tasks, rather than coercing them into a larger one, likely multiplying inference costs. 

\ourname can be used in a continual setting, where at each time $t$ one has a different target $T_t$, and the auxiliary task $A$ is composed of the union of all prior targets $T_0, \dots, T_t$. The retrieval system should automatically determine what information from whatever past targets is relevant to the present, and what information is redundant in $A$ and can be eliminated. The important difference compared to ordinary continual learning is that each step starts with the base model, so there is no catastrophic forgetting, and what is updated is the auxiliary task. In other words, the integration of information occurs in $A$, not in the trained model $f$.

\subsection{Related problems}

\ourname relates to unsupervised domain adaptation (UDA) \cite{MCC, CAN, domainnet}, since the target dataset is not annotated. However, in UDA one assumes that the source dataset $S$ is available along with the target $T$, which is not necessarily the case in \ourname since users may want to bypass annotation altogether, and directly adapt the pre-trained model using the auxiliary dataset $A$, based on the target task $T$, without having direct access to $S$. 

\ourname also relates to semi-supervised learning (SSL) \cite{pseudo_labels, fixmatch, comatch}, since the target dataset $T$ and the auxiliary dataset $A$ are not annotated. However, in SSL one assumes that labeled $S$ and unlabeled data are drawn from the same joint distribution, which is not the case for $T$ and $A$ in \ourname, and, in any case we do not aim to infer labels of $A$, and just use it to improve the model on the target task. 

\ourname is also related to open-set domain adaptation \cite{open_set_DA, open_set_DA_backpropagation}, since the auxiliary dataset $A$ is heterogeneous and does not share the same label space as the source and target task. It is also related to out-of-distribution detection (OOD) \cite{OOD_survey, VOS}, since one needs to decide whether to add samples from the auxiliary dataset, and to active learning \cite{active_learning_survey}, since one needs to decide what samples to add. 

Naturally, \ourname closely relates to test-time adaptation (TTA) \cite{adaContrast, tent, test_time_training, MEMO_test_time_adaptation, SHOT}, and to memory-augmented or retrieval-based architectures \cite{retrieval_augmented_classification, retrieval_diffusion_models, retrieval_generative_models}, widely developed in the language domain \cite{RETRO, retrieval_language_model, retrieval_NLP}, where the hypotheses live in the same space of the data and nuisance variability is limited to paraphrasing.

In summary, \ourname lies at the intersection of UDA, SSL, OOD, TTA, Active Learning, and Retrieval, yet it does not fit neatly into any of them, making both the survey of related literature (Sect.~\ref{sec:related}) and experimental assessment (Sect.~\ref{sec:experiments}) non-straightforward.

\subsection{Key ideas and contributions} 

We propose a method to solve \ournamenospace, based on a target unlabeled dataset $T$, that selects samples from an auxiliary dataset $A$, using a retrieval model $R$.

Starting from any model $f_S$ pre-trained by the provider on a dataset $D$ and later fine-tuned by the user on a labelled dataset $S$, our method finds subsets of an auxiliary dataset $A$ that are relevant for the target dataset $T$, using nearest neighbors in $A$ to samples in $T$, measured in a representation space computed by a retrieval model $R$ (in our case, a CLIP embedding \cite{CLIP}). 

The key technical contribution is a contrastive loss used for updating the model $f_S$ to a new model $f_{A|T}$, whereby negative pairs are selected by retrieving samples from the external dataset $A$ that are {\em informative} of $T$ using the retriever $R$. Furthermore, to improve training stability, we exclude same-class negatives pairs from $T$ by exploiting assigned pseudo-labels obtained by averaging predicted logits on different data augmentations. 
Our method can be thought of as a form of contrastive ``dataset augmentation'' by enlarging the user data with samples drawn from a different (unlabeled) dataset $A$, based on guidance provided by a retriever $R$. This procedure can be followed by both the user and the provider, thus empowering them to adapt the core model (train-time adaptation) or a sequence of disjoint custom models (test-time adaptation). 

We show that applying \ourname improves downstream classification accuracy over the paragon supervised fine-tuning \cite{LQF, model_zoo} for train-time and test-time adaptation methods \cite{tent, SHOT, adaContrast} for test-time. In particular, as the number of data available during adaptation decreases, \ourname improves by up to 13\% and 8\% in relative Top1 accuracy at train and test time, respectively.

\begin{figure*}
    \centering
    \includegraphics[width=0.8\linewidth]{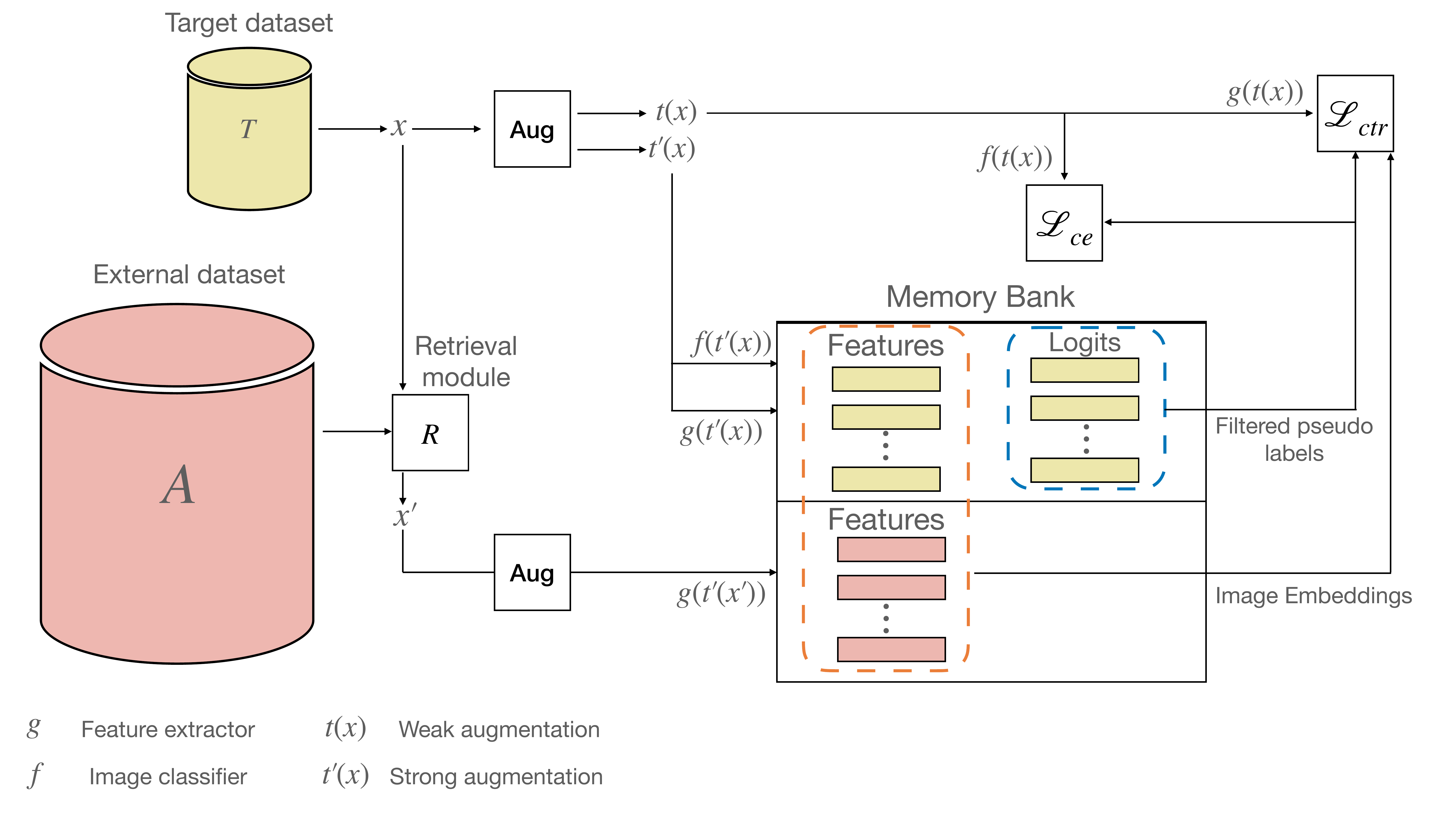}
        \vspace{-0.3cm}
    \caption{\textbf{Our proposed framework.} Given an image $x$ from the unlabeled dataset $T$, an auxiliary external dataset $A$, and a retrieval method $R$, \ourname strongly augments $x$ with $t^\prime$ and stores the logits and features in a memory bank. Moreover, for each image $x$ in $T$ an independent retrieval system $R$ (\eg CLIP) retrieves from an external data pool $A$ a given number of {\em related} images that are strongly augmented and saved as features into the same memory bank (logits of retrieved images are discarded). At each iteration, for each image $x$ ``filtered'' pseudo-labels are generated leveraging logits in the memory bank. Pseudo-labels are then used both as targets to train the class predictor on weakly-augmented images with $t^\prime$ and as supervision for the contrastive loss. The contrastive loss is computed following instance discrimination on the augmented views of the same image against embeddings of images with different pseudo-labels and embeddings retrieved from the auxiliary pool of data $A$.}
    \label{fig:retrieved_nearest_neighbours}
\end{figure*}

\section{Related work}\label{sec:related}

As we anticipated in the introduction, the problem we tackle has close connections with a number of areas of investigation in machine learning, including UDA, SSL, OOD, TTA, Active Learning, and Retrieval.%

\paragraph{UDA and TTA} 
Unsupervised Domain Adaptation (UDA) has a long history and it has been explored in a variety of different visual tasks, image classification \cite{CAN, MCC, domainnet}, object detection \cite{UDA_OD} and semantic segmentation \cite{UDA_semantic_seg}. 
The main goal of UDA methods is to reduce the performance drop of pre-trained models when deployed on shifted target domains without using any target annotation. One of the most successful ideas in UDA literature is source and target feature space alignment. For example, \cite{UDA_MJ} exploits Maximum Mean Discrepancy, \cite{domainnet} leverages a multi-source moment matching objective, \cite{MCC} uses a non adversarial reduction of the class confusion and \cite{CAN} employs a contrastive adaptation objective to model intra-class and inter-class domain discrepancy.
However, all these methods require knowledge of the target distribution before model deployment, which highly limits their applicability in the wild. 
On the other hand, typical test-time adaptation (TTA) methods only use the target dataset during adaptation \cite{adaContrast, tent} and usually no modification to the pre-training loss is allowed (a notable exception is \cite{test_time_training}). Therefore, test-time adaptation is carried out exploiting regularities between source $S$ and target data $T$. For example, it is often assumed that the target data shares the same class distribution with the source one, or that the un-adapted decision function is not far from the target \cite{tent}. 
Under these assumptions, \cite{tent} minimizes the entropy of the predictions to quickly adapt a given pre-trained model. \cite{MEMO_test_time_adaptation} takes this approach one step further and exploits different synthetic data augmentations to further improve performance. 
Among other test-time adaptation methods, AdaContrast \cite{adaContrast} is the closest to our solution since it leverages a contrastive loss for adaptation. However, as in previous methods, only synthetic data augmentations are used to construct the self-supervised contrastive loss. 
On the other hand, our method is not bounded to synthetic data augmentations and augments samples in $T$ by leveraging other real data to better capture the variability in $T$. 

While \ourname is close to TTA \cite{adaContrast, tent, SHOT, test_time_training}, it differs in that we expect that the dataset used for adaptation is not just $T$, which is assumed to share the same label space of $S$, but also $A$, a typically very large dataset largely irrelevant to $T$. Hence, we leverage a retriever $R$ to find the needles in the haystack, an element not present in the TTA literature.

\paragraph{Retrieval/memory augmented models} 
Recently, retrieval based models have been used to solve symbolic manipulation \cite{Neural_Turing_Machine}, anomaly detection \cite{anomaly_detection_retrieval}, image generation \cite{retrieval_diffusion_models} and image  classification \cite{retrieval_augmented_classification}. 
In particular, \cite{retrieval_augmented_classification} shows that augmenting a standard image classification model with an explicit image retrieval module highly improves accuracy on long tailed classification datasets. \cite{retrieval_diffusion_models}, instead, uses retrieved images as guidance for generating highly detailed uncommon concepts. 
Retrieval based models have also found applications on other domains other than Computer Vision.
For example, in the NLP domain, several recent methods leverage large corpora to augment pre-trained large language models predictions with a non-parametric memory module \cite{RETRO, retrieval_language_model, retrieval_NLP, retrieval_prompt}. 
In particular, \cite{RETRO} shows that augmenting a large pre-trained language model with an external indexable database has mainly one advantage: higher performance \wrt the number of deployed parameters, which in turn unlocks the use of smaller/faster models that are less likely to memorize the training data.
However, this result has yet to be reliably verified for large scale computer vision models. One of the main reasons for this discrepancy is that in the language domain the query and the data/representation live in the same space, so the answer to a query, expressed as a string of text,  is a string of text which may potentially be in the knowledge base or easily interpolated from it. However, in the image domain it is usually not reasonable to assume that the answer to a given query already exists in some indexable database or knowledge base (\eg downstream labels might differ from pre-specified labels in the knowledge base or database). Hence, in our case, we do not assume the auxiliary dataset $A$ has ready answers to our queries.

\section{Method}

We assume that there is a {\em provider} who pre-trains a model $g$ on a dataset $D$ obtaining $g_D: X \rightarrow Z$ where $X$ are RGB images and $Z = {\mathbb R}^{d}$ where $d$ is the dimension of the feature space. Here, $D$ is a large dataset which is typically not accessible after pre-training and may, with time, become obsolete.

A {\em user} has access to $g_D$, but wishes to improve it on a specific dataset $S$ to build a custom classifier $f_S: X \rightarrow Y$, using $g_D$ as a backbone, and fine-tuning it along with a linear layer. 

Test data owned by the user and optionally made available to the provider, is drawn from an unlabeled dataset $T$, which may be different from both $S$ and $D$, but shares the same hypothesis space $Y$ of $S$  \cite{adaContrast, tent, test_time_training}. In particular, there may be a domain shift from $S$ to $T$, or the two may pertain to entities, such as products or fashion, that evolve over time. If such out-of-distribution phenomenon is manifest in the test samples, not only $g_D$, but even $f_S$ will perform poorly.

The goal of \ourname is to train an adapted model, leveraging an auxiliary dataset $A$, starting from $f_S$, but without directly accessing $S$, and leveraging instead unlabeled data from $T$ available at inference time. We call the resulting adapted model $f_{A|T}$, where the dependency on $S$ is implicit in its pre-training.
We note that $A$ may be a private dataset, accessible to the user but not the provider. Conversely, the provider may have an internal dataset that may be available to adapt the model to commonly observed tasks from opted-in users, using the same process followed by the user to perform test-time adaptation.

The goal of \ourname is to train a model $f_{A|T}$ that improves the baseline $f_S$ and gets as close as possible to the paragon which is to train with the entirety of the datasets $D, S, A$ and $T$. To this end, we consider any generic pre-training $g_D$ and fine-tuning $f_S$, and perform {\bf retrieval} by finding subsets of $A$ that are informative of $T$. We do so by finding nearest neighbors of samples of $T$ in $A$ using CLIP embedding space computed by a retrieval model $R$. We then perform {\bf aggregation} to combine $T$ with $A$ into an ``augmented training set''. This is the key technical contribution of our work and is implemented as follows. 
Given each datum in $x \in T$, we create multiple augmentations $x_i$, and use $f_S$ to compute the corresponding pseudo-labels.  
Then, we consider a contrastive loss whose objective is to pull closer features of different views (positive pairs) while pushing away features of different images (negative pairs). We consider as negatives the retrieved samples from $A$ that are neighbors to $T$ with $R$ and samples in $T$ with negative pseudo-labels. 

\ourname finds samples in $A$ that are synergistic with $T$ as contrastive neighbors and avoids pushing away same-class pairs to learn better semantically meaningful clusters. Redundant information is avoided simply by removing samples with near-duplicate embeddings according to $R$ \cite{RETRO, retrieval_generative_models}.
        
Note that the contrastive loss does not update the linear classifier but only the features. So, \ourname updates the classifier $f_{A|T}$ with supervision from pseudo-labels generated from $T$ \cite{fixmatch, comatch, adaContrast}. 

We now describe each component of method in detail.

\paragraph{Retrieval module}
Since the search cost scales with the data pool size any slow retrieval algorithm is not a feasible solution \cite{yan2020neural, thinking_fast_and_slow}. We therefore use a fast retriever $R$ whose main goal is to filter irrelevant data in $A$ given target samples in $T$ (\eg out-of-distribution or near duplicate).

The retrieval module consists of a general image encoder $R : X \to \mathbb{R}^d$, that we use to index the auxiliary pool of images $A$. We note that different retrieval systems lead to largely different retrieval distributions that mostly depend on the invariance classes imposed during the retrieval pre-training objective. For example, a CLIP model \cite{CLIP} is trained to match images with likely captions, while a self-supervised model (\eg DINO \cite{dino}) is trained to be invariant to per-sample synthetic data augmentations. We evaluate the impact of the retrieval choice in \cref{tab:CLIP_vs_DINO_vs_random_retrievals}.
Differently from \cite{adaContrast, RETRO}, we do not use specialized fast approximate nearest-neighbor search (such as FAISS \cite{faiss} or SCaNN \cite{scann}). Instead, we simply use a brute force search on the most similar keys (embedding indexes), since when the size of the external database does not exceed $10M$ the time reduction of approximate nearest neighbor search is minor.

\paragraph{Encoder initialization}
Since \ourname does not impose any restriction on the objective used to pre-train $g_D$, we shall consider models pre-trained both with a supervised \cite{model_zoo} or self-supervised \cite{dino} objective (see \cref{tab:train_adaptation_comparison}).

\subsection{Learning objective}
Our learning objective consists of two parts. First, a self-supervised objective function that is used to incorporate retrieved information from $A$ both at train and test time (see \cref{sec:retrieval_contrastive_loss}). Second, a cross-entropy objective that is driven by ground truth labels at training time and by pseudo-labels at test time (\cref{sec:pseudo_labeling_consistency_regularization}). 

\subsubsection{Retrieval-augmented objective function}\label{sec:retrieval_contrastive_loss}
In this section we describe the self-supervised objective function that we exploit to incorporate information from $A$.
Inspired by recent advances in self-supervised objective designs \cite{simclr, MoCo, dino} we exploit a contrastive objective driven by pairwise information. In particular, we follow the \textit{instance-discrimination principle}: features of different views of the same image (positive pairs) are pulled closer, while features of different images (negative pairs) are pushed away. 
The key insight is that, even in presence of domain shift, the contrastive loss discriminative power increases with more negative samples \cite{supcon, simclr}. However, adding easily separable negatives does not provide much learning signal. We therefore use the retrieval module $R$ to modify the noise distribution and gather images that serve as \textit{harder negatives}. 

\paragraph{Retrieval-augmented contrastive loss}

As in \cite{adaContrast}, given an image $x$, we create a weakly augmented view $t(x)$ and a strongly augmented view $t^\prime(x)$. 
Then, we apply the InfoNCE loss on $q=g(t(x))$, $k=g(t^\prime(x))$ and the set of strongly augmented negatives $\mathcal{N}_q$. Here, $g$ denotes the last layer features (before the classifier head) extracted by the model $f_{A|T}$ being trained and the set $\mathcal{N}_q \subset A \cup T$ is composed of different-class samples from $T$ and by retrieved samples from $A$ (nearest neighbors of $x$ according to $R$).    

\begin{equation}\label{eqn:contrastive_loss}
    \mathcal{L}_{\text{ctr}}(x) = - \log \frac{\exp(q \cdot k / \tau)}{\sum_{j \in \mathcal{N}_q}  \exp{ (q \cdot k_j / \tau)}}
\end{equation}
where $\tau$ is a temperature hyper-parameter and all $k_j$ are feature embeddings stored in a memory bank of length $n_p$ that is updated by appending the new embedding $k$ at each step \cite{MoCo, adaContrast}.

As observed in \cite{supcon, simclr, adaContrast, MoCo}, the InfoNCE loss in \cref{eqn:contrastive_loss} might strive to minimize the cosine distance between $q$ and $k$ while maximizing the cosine distance of $q$ and all the negatives in the denominator. 
In particular, not pushing away same-class pairs helps in building a feature space that is more aligned with the semantic of the downstream task. Therefore, when the label information (or pseudo-labels) is available, we modify $\mathcal{N}_q$ not to include samples with the same label $y$ (or pseudo-label $\bar{y}$) of $x$: 
\begin{equation}\label{eqn:neighbourset_labels}
    \mathcal{N}_q^{\text{lab}} := \{j \mid y \neq y^j\} \cup \emptyset   
\end{equation}
In \cref{sec:pseudo_labeling_consistency_regularization} we describe how to compute pseudo-labels $\bar y$ on the target set $T$.

Furthermore, we leverage the auxiliary data available in $A$ to increase the number of negatives. However, as observed in \cite{divide_and_contrast} naively leveraging a large pool of uncurated data in a self-supervised contrastive loss might not lead to performance improvements since negatives can be less informative (easy negatives). To overcome this limitation we leverage the retrieval system $R$ whose task is to gather more relevant negatives (\textit{hard negatives}). 
More specifically, we build $\mathcal{N}_q^{\text{ret}}:=\text{NN}_{n_r}(q)$ as the set of $n_r$ nearest neighbors of $x$ from $A$. 
Note that only retrieving from the nearest neighbors might be counter-productive, since many nearly duplicate images could be retrieved and considered as negatives. 
This phenomenon gets sharpened if there is small/no distribution shift between adaptation data and the external pool of samples. \ourname solves this with a simple deduplication strategy applied to the retrieved data.
We propose to randomly extract $k$ samples among $\text{NN}_{r \times n_r}(q)$, \ie we first select the top $r\cdot n_k$ retrieved data, and then uniformly sample a subset of $k$ samples. In this way, even if the topmost retrieved samples are near duplicates to the query image the likelihood of treating them as negatives is reduced. In our experiments we find that $r=5$ is a robust choice across different experiments.

To conclude, the set of negative examples we use in \cref{eqn:contrastive_loss} is $\mathcal{N}_q = \mathcal{N}_q^{\text{lab}} \cup  \mathcal{N}_q^{\text{ret}}$.

\paragraph{Ground truth labels vs pseudo-labels}
In \ourname it is possible to adapt pre-trained models not only at test-time (by the user) but also at train time (by the service provider) as new data become available. In the latter case, ground truth labels might be available and should not be discarded. 
Our method can be modified to work with ground truth labels by incorporating them into its main objective in place of pseudo-labels so that ground truth labels are used to avoid same-class negatives in \cref{eqn:contrastive_loss} and are used to directly supervise the model predictions. 
On the other hand, at test-time (when ground truth labels are not available), \ourname exploits the close-set assumption and uses pseudo-labels \cite{comatch, adaContrast}.  
However, the quality of pseudo-labels is important, in  \cref{sec:pseudo_labeling_consistency_regularization} we propose a simple refinement strategy to get higher quality ``filtered'' ones.

\subsubsection{Supervised/weakly-supervised objective}\label{sec:pseudo_labeling_consistency_regularization}
Since the contrastive loss does not update the linear classifier but only the features of the predictive model, we incorporate supervision into the objective function by exploiting labels $y$ (if available) or, more generally, pseudo-labels $\bar y$, that are generated by the hypothesis $f_{A|T}$  \cite{fixmatch, comatch, adaContrast}. However, pseudo-labels are known to be noisy, especially if $T$ is different from $S$ \cite{adaContrast, comatch}, therefore we propose to further refine them by leveraging other augmented views of the same image $x$ \cite{MEMO_test_time_adaptation}. 
\vspace{-0.15cm}
\begin{equation}\label{eqn:pseudo-labels}
\vspace{-0.15cm}
    \bar{y}(x) = \arg\max_i \bar f_{A|T}(x)
\end{equation}
where  $\bar f_{A|T}(x)$ is obtained by averaging logits with respect to strong synthetic data augmentations of $x$. To improve efficiency of our method and reduce training time \cite{adaContrast, comatch}, we implement this exploiting a memory bank which contains past predicted logits and features (see  \cref{sec:retrieval_contrastive_loss}).

We note that using ``filtered'' pseudo-labels to guide model adaptation can be interpreted as a form of consistency regularization or distillation, which, in the case of semi-supervised learning, has the main objective of propagating known labels towards unlabelled samples \cite{fixmatch, comatch}. Overall, our supervised loss/consistency regularization is implemented as:
\begin{equation}\label{eqn:cross-entropy-consistency-regularization}
    \mathcal{L}_{\text{ce}}(x) = \mathbb{E}_{x \in \mathcal{D}_t} H(\bar{y}(x), f_{A|T}(t(x)))
\end{equation}
where $H(a, b) = -\sum_{c=1}^C a_c \log b_c$ and $\bar{y}(x)$ is the ``filtered'' pseudo-label.

\begin{table}[h!]
\centering
\caption{\textbf{Comparison with transfer learning baselines.} Classification Top1 Accuracy (\%) on fine-grained downstream datasets. Bold is the highest.  Comparison of \ourname with supervised transfer learning (fine-tuning) on ResNet50. We show that \ourname performs on par with a strong supervised fine-tuning baseline on high shot fine-grained tasks. Moreover, when the number of samples allowed during adaptation is reduced (20\% of the original datasets) we show that the use of an external data pool of images allows \ourname to  perform better on different fine-grained tasks.}
\resizebox{\columnwidth}{!}{
\begin{tabular}{c|cc | cc|cc | cc} \toprule
                  & \multicolumn{4}{c|}{20\% of samples}    & \multicolumn{4}{c}{100\% of samples}\\ 
                  & \multicolumn{2}{c}{Sup.\@ } & \multicolumn{2}{c|}{Self Sup.\@ }     & \multicolumn{2}{c}{Sup.\@ }           &  \multicolumn{2}{c}{Self Sup.\@ }\\ 
     {Dataset}    & {Sup.\@ FT} & \ourname & {Sup.\@ FT} & \ourname &{Sup.\@ FT} & \ourname & {Sup.\@ FT} & \ourname \\ \midrule 
    Stanford Cars &  61.4       & \textbf{66.0}  &   31.7      & 64.6  & \textbf{93.5}   & \textbf{93.5} & 93.2 & 93.0 \\
    Aircrafts     &  11.8       & 35.0  &   39.9      & \textbf{60.0}  & 86.4   & 88.4 & 88.2 & \textbf{89.1} \\
    CUB200        &  52.0       & \textbf{55.5}  &   27.7      & 43.6  & 82.2   & \textbf{82.4} & 80.0 & 80.3 \\
    MIT-67        &  60.9       & \textbf{67.6}  &   62.8      & 66.4  & 77.2   & \textbf{77.6} & 76.8 & 75.9 \\
    Stanford Dogs &  86.8       & \textbf{87.3}  &   40.9      & 56.5  & \textbf{92.2}   & 89.6 & 76.5 & 81.9 \\
    \bottomrule
\end{tabular}
}
\label{tab:train_adaptation_comparison}
\end{table}

\section{Experiments}\label{sec:experiments}

\subsection{Experimental setup}\label{sec:experimental_setup}
We evaluate \ourname on standard train and test time adaptation benchmarks. At train-time, \ourname is applied on fine-grained classification datasets as done in \cite{model_zoo, LQF}. In particular, we use 
MIT-67 \cite{mit67recognizing}, 
CUB-200 \cite{WahCUB_200_2011}, 
FGVC-Aircraft \cite{maji13fine-grained}, 
Stanford Cars \cite{stanfordcars}, 
Stanford Dogs \cite{stanforddogs}. 
At test-time, following \cite{adaContrast}, we use a closed set benchmark composed of VisDA-C \cite{VISDA} and DomainNet-126 \cite{domainnet} (we use DomainNet-126 and not DomainNet since the latter has noisy labels \cite{adaContrast}). DomainNet-126 contains 126 concepts shared across four domains (Real, Sketch, Clipart, Painting), while VisDA-C is a 12 class dataset that focuses on synthetic-to-real adaptation. 
To build the large pool of external data $A$ we use images (without labels) from the following datasets: ImageNet1k \cite{imagenet_cvpr09}, iNaturalist 2019 \cite{iNaturalist_HornASSAPB17}, Food-101 \cite{food101}, Logo 2k+ \cite{logo2k}, NWPU-RESISC 45 \cite{nwpu_resisc45}, iMaterialist Product \cite{imaterialist}. Overall, size of the auxiliary dataset $A$ is $\approx 2M$ images aggregated from a range of different domains and applications.

\paragraph{Baselines} 
For test-time adaptation we compare our method with both Unsupervised Domain Adaptation and Test-Time Adaptation methods. For UDA methods, we compare to CAN \cite{CAN} and MCC \cite{MCC} since they have been reported to be the best performing methods on our chosen benchmarks. For TTA we compare with TENT \cite{tent}, SHOT \cite{SHOT} and AdaContrast \cite{adaContrast}. We do not directly compare with TTT \cite{test_time_training} since it requires to modify the pre-training objective function and is therefore not a truly test-time only adaptation method. 
All the baseline results on train-time training are obtained following supervised fine-tuning best practices (\eg data augmentations such as MixUp \cite{mixup} and RandAugment \cite{randaugment}, linear warmup and cosine annealing learning rate schedules \cite{ViT}) and running extensive hyper-parameter search (see \cref{sec:hps_appendix} for details). 
Similarly to previous works \cite{adaContrast, LQF, model_zoo}, we use ConvNets architectures (ResNet50/101) pre-trained using both supervised \cite{model_zoo} and self-supervised \cite{dino} objective functions on ImageNet1k.

\subsection{Train time model adaptation with retrieval}

At train-time \ourname takes as input a model pre-trained on some pre-training data (either with supervision or self-supervision), a labelled dataset $S$ and large database of images $A$ and adapts the pre-trained features to the downstream task. Performance is evaluated on held out data $T$ that is not used for further adaptation. This mimics the typical model customization scenario (transfer learning \cite{model_zoo, LQF}) solved with supervised fine-tuning. In this section, we pick $S$ to be a labelled fine-grained classification dataset from the ones listed in \cref{sec:experimental_setup}.

In \cref{tab:train_adaptation_comparison} we test how much retrieving samples from $A$ help \ourname at training time.
We compare \ourname adaptation against supervised fine-tuning of two different pre-trained backbone models both in the high and low data regime (see \cref{sec:experiments_design_appendix} for details on datasets subsampling). 
Therefore, we either use the whole downstream dataset (100\%) or we subsample it by keeping only $20\%$ of the labelled training data (all the remaining data are discarded, and no further used). 
Our models are pre-trained with a supervised objective or a self-supervised one on ImageNet1k. 
Note that \ournamenospace, compared to the baselines, improves feature adaptation in both data regimes and it is effective regardless of the backbone choice. 
In particular, supervised pre-trained features improve 13\%/5\% while self-supervised 30\%/4\% on the low and high data regime respectively. 
Our results show that $A$ can be leveraged to add relevant information during adaptation even if the external data come from a different distribution.
We further study the effect of adding more retrieved samples in \cref{sec:ablations}, our results suggest that increasing the number of retrieved images saturates relatively early and the trade-off between computational cost (the more the retrievals the higher the training time) and performance is relatively stable across different datasets. In particular, the performance starts saturating as soon as the retrieved dataset is twice as large as the training dataset \cref{fig:T3AR_ablation_num_retrievals} in the appendix.

\begin{table}[t!]
\centering
\caption{\textbf{Comparison with UDA and TTA baselines.} Avg. Classification accuracy (\%) on 7 domain shifts of DomainNet-126 and on 1 domain shifts of VisDA-C train $\to$ val for different target $T$ dataset sizes (1\%, 10\% and 100\%). 
Bold is the highest. 
\ourname achieves the highest average performance when few samples are available for adaptation, $1\%$ and $10\%$ of the whole dataset.
}
\resizebox{\columnwidth}{!}{
\begin{tabular}{c|ccc|ccc}
    \toprule
    Method & \multicolumn{3}{c}{DomainNet-126} & \multicolumn{3}{c}{VisDA-C} \\
                   & 1\% &  10\% & 100\% & 1\% &  10\% & 100\% \\
    \hline
    CAN \cite{CAN} &    - & - & -    & - & - & \textbf{87.2} \\
    MCC~\cite{MCC} &    - & - & 48.9 & - & - & 78.8 \\
    \midrule
    Source only       & 55.6 & 55.6 & 55.6 & 43.8 & 43.8 & 43.8 \\
    TENT~\cite{tent}  & 53.7 & 54.0 & 57.7 & 45.7 & 46.9 & 49.2 \\
    SHOT \cite{SHOT}  & 57.2 & 64.1 & 67.1 & 63.6 & 69.1 & 83.0 \\
    AdaContrast \cite{adaContrast}  & 60.6 & 65.8 &  \textbf{67.8} & 68.3 & 72.8 & \textbf{87.2}\\
    \ourname   & \textbf{63.5} & \textbf{66.3} & 67.5 & \textbf{70.2} & \textbf{77.5} & 85.9\\
    \bottomrule
\end{tabular}
}
\label{tab:test_time_adaptation_for_different_datasetsizes}
    \vspace{-0.4cm}
\end{table}

\begin{table}[h!]
\centering
\caption{\textbf{Do we need an expert retrieval module?} We compare train-time downstream Classification Top1 Accuracy (\%) of \ourmethod on fine-grained classification tasks when equipped with random or expert retrieval module (\eg CLIP, DINO). Even a non expert retrieval system does not jeopardize generalization. Nonetheless, the average relative performance drop \wrt to expert retrieval systems is $\approx 25\%$. And the stronger CLIP retrieval leads to better results.}
\begin{tabular}{ccccc} \toprule
    {Dataset}     & Random $R$ & DINO & CLIP & \\ \midrule
    Stanford Cars &  61.4  & 62.4 & \textbf{66.0}   \\
    Aircrafts     &  18.4  & 31.6 & \textbf{35.0}   \\
    CUB200        &  48.2  & 54.0 & \textbf{55.5}   \\
    MIT-67        &  62.2  & 66.6 & \textbf{67.6}   \\
    Stanford Dogs &  83.9  & 86.9 & \textbf{87.3}   \\
        \bottomrule
\end{tabular}
\label{tab:CLIP_vs_DINO_vs_random_retrievals}
\vspace{-0.4cm}
\end{table}

\subsection{Test time model adaptation with retrieval}
At test-time \ourname takes as input a model pre-trained on the source dataset $S$ whose labels space is the same as the one in the unlabelled target set $T$. However, the distribution of images in $S$ need not be the same as in $T$ (covariate shift). The performance of our method is evaluated on the average Top1 accuracy on different domains (7 for DomainNet-126 and 1 for VisDA-C). As in previous experiments, the auxiliary data pool $A$ is taken as the concatenation of the datasets listed in \cref{sec:experimental_setup}. To compare our results with TTA literature \cite{adaContrast, tent, SHOT}, and only in this experiment, we fix the pre-trained backbones as the ones used in \cite{adaContrast}. More specifically, we add a 256-dimensional bottleneck consisting of a fully-connected layer followed by a BatchNorm layer after the backbone, and apply WeightNorm on the classifier, for more details we refer to \cite{adaContrast}.

Previous results in the literature \cite{adaContrast, tent, SHOT, test_time_training} assume that \textit{all target data} $T$ are used for adaptation. However, relying on plenty of samples for adaptation, even if unlabelled, could be a limiting factor in many real world scenarios. 
In \cref{tab:test_time_adaptation_for_different_datasetsizes} we test the capability of \ourmethod to efficiently adapt when little target data are available (1\%, 10\% and 100\% of $T$). 
\ourname achieves high Top1 average accuracy both on DomainNet-126 and VisDA-C benchmarks. In particular, the fewer the data available at test time the higher the performance gap \wrt other state of the art methods.
As in the train-time experiment, we observe that the retrieval system plays an important role. In fact, while other methods \cite{adaContrast, tent, SHOT} mainly rely on synthetic data augmentations to compensate for the lack of target data, our method also leverages retrieved real images that enable the learned features to better approximate the target data manifold.

\subsection{Ablation studies}\label{sec:ablations}

\paragraph{Do we need an expert retriever $R$? And does distribution shift of $A$ \wrt $T$ impact performance?} 
Intuitively, the performance of \ourname could be upper bounded by the Top-1 accuracy of its retrieval system. And the higher the domain gap of the retrieved samples from $A$ \wrt target data in $T$ the worse the downstream performance gets.

In \cref{tab:CLIP_vs_DINO_vs_random_retrievals} we answer the first question by comparing a random retrievals, and two expert retrieval systems, one based on DINO \cite{dino} and the other based on CLIP pre-training \cite{CLIP}. Both DINO and CLIP embeddings achieve high performance on zero-shot classification on the fine-grained datasets we use (see \cref{sec:retrieval_accuracy_appendix}). 
In particular, in \cref{tab:CLIP_vs_DINO_vs_random_retrievals} we show that even a non-expert retrieval system does not completely jeopardize generalization, the average relative performance drop \wrt to an expert retrieval system is 20/25\% and, for some datasets, it is comparable with the supervised fine-tuning results in \cref{tab:train_adaptation_comparison}. 
This observation suggests that even randomly retrieved images can act as a generic regularizer and do not harm generalization.

In \cref{fig:domain_gap_on_the_external_data_pool} we answer the second question by artificially introducing a controlled distribution shift on the retrieved samples. In particular, we progressively include more data from the adaptation domain in the external pool that, in turn, are more likely to be retrieved by $R$. 
We note that the higher the domain gap is the lower the final accuracy gets, since finding hard (informative) negatives becomes harder.

\begin{figure}
    \centering
    \includegraphics[width=0.9\linewidth]{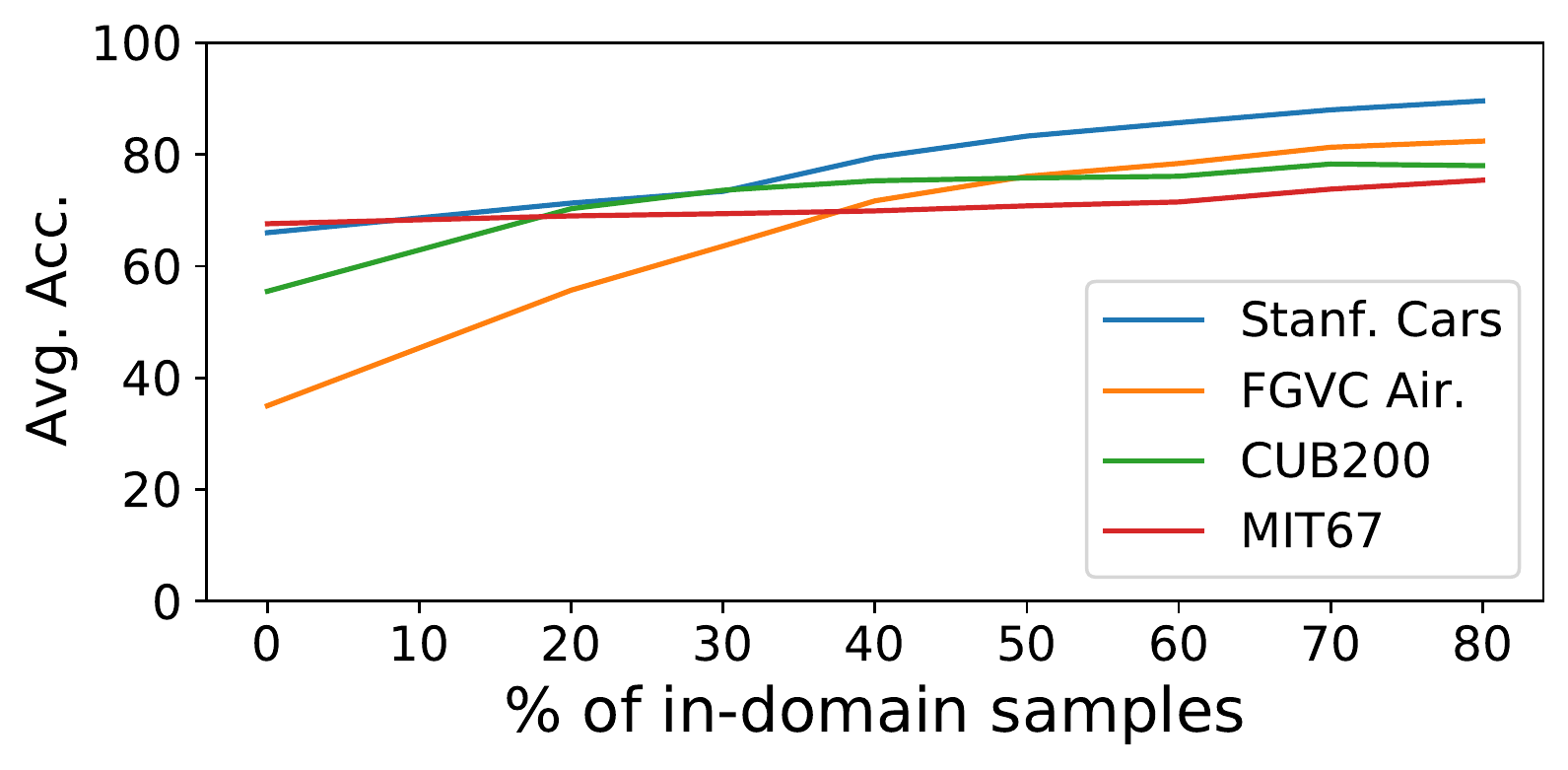}
    \caption{\textbf{How much distribution shift can \ourmethod tolerate?} We compare \ourname in a train time setting on fine-grained classification datasets as the domain gap between the adaptation data and the auxiliary data $A$ increases. To artificially control the distribution shift we progressively include more adaptation data, which are more likely to be retrieved, to the external pool. The higher the domain gap the lower the performance.}
    \label{fig:domain_gap_on_the_external_data_pool}
    \vspace{-0.5cm}
\end{figure}

\textbf{Impact of the size of $A$.} We test the sensitivity of \ourname to the size of the external set by using a $10\%$ subset of $A$, ImageNet-21k and its subset ImageNet-1k. In \cref{tab:A_ablations}, we show that increasing the size of the external data pool leads to higher average accuracy. However, using larger datasets is not as helpful as having better domain coverage. The gaps are 0.4\% (Subset-$A$ $\to$ $A$), 0.7\% (Subset-IN21k $\to$ IN21K), 1.2\% (Subset-IN21k $\to$ $A$), 0.5\% (IN21k $\to$ $A$). Interestingly, differently from \cite{quality_not_quantity_Nguyen} no task competition is present in \ournamenospace. In fact, thanks to the retrieval module that ignores what is not relevant, increasing the external dataset size strictly improves results. 

\textbf{Impact of the domain coverage of $A$.}
We replace our external dataset of 2M images with a 2M random subset of ImageNet-21k. In \cref{tab:A_ablations} we show that \ourname still improves over the baselines of not using an external data pool ($54.6\% \to  62.0\%$ for train-time adaptation and $69.3\% \to 70.6\%$ for test-time adaptation). However, our choice of $A$  has better overall results ($62.3\%$ and $71.9\%$ for train and test time respectively) due to better domain coverage.

\textbf{Ablation over the composition of $A$.} In \cref{tab:A_ablations} we ablate over the datasets used to build $A$. Removing ImageNet-1k from the external pool leads to 7.5 \% average drop in performance, while dropping iNaturalist or Logo 2k is not as harmful and the average gap is $\approx 1\%$.
In particular, we found that IN-1k (the largest dataset in $A$) provides most of the retrieved samples (more than 85\%) both during Train- and Test-Time adaptation. 
However, there are some exceptions: CUB200 retrieves half of the data from iNaturalist, while DomainNet-126 (on all domains) retrieves more than 15\% samples from Logo-2k and 5\% from iNaturalist.
\paragraph{Sensitivity to the number of retrievals}
In \cref{sec:detailed_results_appendix} in the appendix we study the sensitivity of \ourname to the number of allowed retrieved images. Our results across different datasets show a diminishing return in performance as the number of NNs increases (see \cref{fig:T3AR_ablation_num_retrievals}). 
Since retrieving more samples increases (linearly) adaptation time, our experiments suggest that a trade-off, to discount compute over marginal accuracy improvements, is to retrieve no more than twice as many samples as the target dataset.

\begin{table}
\centering
\caption{\textbf{Ablations on the external dataset.} Accuracy on downstream tasks (rows) when using different external datasets (columns). Results are reported on the same 20\% subsets used for train time experiments (see \cref{tab:train_adaptation_comparison}) and 10\% subsets used for test time experiments (see \cref{tab:test_time_adaptation_for_different_datasetsizes}). By $A$-IN1k, $A$-iNat, $A$-Logo we denote $A$ ablated of the corresponding dataset.
}
\resizebox{\columnwidth}{!}{
\begin{tabular}{c|c|cc|c|cc||c|c|c}
    \toprule
     &&  \multicolumn{2}{c|}{$A$} & \multicolumn{1}{c|}{IN1k} & \multicolumn{2}{c||}{IN21k} & \multicolumn{1}{c|}{$A$-IN1k} & \multicolumn{1}{c|}{$A$-iNat} & \multicolumn{1}{c}{$A$-Logo} \\
      &     SOTA      & 100\% & 10\% & 100\% & 15\% &  100\% & & &  \\
    \hline
    Cars & 61.4 & 66.0 & 65.8 & 65.6 & 66.8 & \textbf{67.3} & 60.3 (-5.7)  &  65.2 (-0.8) & 65.2 (-0.8) \\
    Air. & 11.8 & \textbf{35.0} & 33.8 & 33.2 & 31.9 & 32.8 & 17.8 (-17.2) &  34.3 (-0.7) & 34.7 (-0.3) \\
    CUB  & 52.0 & \textbf{55.5}  & 55.1 & 53.7 & 53.7 & 54.5 & 53.4 (-2.1)  &  54.0 (-1.5) & 55.4 (-0.1) \\
    MIT  & 60.9 & 67.6  & 67.5 & 67.1 & 67.7 & \textbf{68.5} & 64.5 (-3.1)  &  67.4 (-0.2) & 67.3 (-0.3) \\
    Dogs & 86.8 & \textbf{87.3} & 87.2 & 87.2 & 86.9 & 86.9 & 82.1 (-5.2)  &  86.8 (-0.5) & 86.9 (-0.4) \\
    \midrule
    Train Avg. & 54.6 & \textbf{62.3} & 61.9 & 61.4 & 61.4 & 62.0 & 55.6 (-6.7) & 61.5 (-0.8) & 61.9 (-0.4)  \\
    \toprule
    \midrule
    DNet-126 & 65.8 & \textbf{66.3} & 65.7 & 63.8 & 64.5 & 64.7 & 57.6 (-8.7) &  63.2 (-3.1) & 62.8 (-3.5) \\
    VisDA-C & 72.8 & \textbf{77.5} & 77.0 & 76.6 & 75.1 & 76.5 & 66.8 (-10.7) &  77.2 (-0.3) & 76.8 (-0.7) \\
    \midrule
    Test Avg. & 69.3 & \textbf{71.9} & 71.4 & 70.2 & 69.8 & 70.6 & 62.2 (-9.7) & 70.2 (-1.7) & 69.8 (-2.1) \\
    \midrule
    \midrule
     Avg. & 58.8 & \textbf{65.0} & 64.6 & 63.9 & 63.8 & 64.5 & 57.5 (-7.5) & 64.0 (-1.0) & 64.2 (-0.8) \\
    \bottomrule
\end{tabular}
}
\label{tab:A_ablations}
\vspace{-0.5cm}
\end{table}

\section{Conclusions}\label{sec:conclusions}

We introduced \ourname to adapt pre-trained models both at train and test time by means of a retrieval module and a searchable pool of auxiliary samples. 
Differently from previously proposed methods \cite{adaContrast, MEMO_test_time_adaptation} that by-pass the lack of a adaptation data by introducing specific self-supervised objectives driven by data augmentations, \ourname builds a self-supervised objective that is driven by real data, thus better capturing the target real data manifold. 
Furthermore, similarly to \cite{adaContrast, comatch}, \ourname exploits ``filtered'' pseudo-labels to align the output distribution of the model to the downstream class labels.
\ourname improves downstream fine-grained classification over standard fine-tuning baselines. Moreover, we compared our method against state of the art test-time adaptation algorithms \cite{adaContrast, tent, SHOT, MEMO_test_time_adaptation} and showed that it resulted in more robust and generalizable features, especially when the available data at test-time are scarce.

{\small
\bibliographystyle{ieee_fullname}
\bibliography{egbib}
}

\clearpage

\renewcommand{\thesection}{\Alph{section}}
\renewcommand{\thesubsection}{\Alph{section}.\arabic{subsection}}
\setcounter{section}{0}

\section*{Supplementary Material}

\section{Implementation details}

\subsection{Architectures}
In our main experiments we use supervised and self-supervised pre-trained ResNets from \cite{timm_repo} and \cite{dino} respectively. 
We implement the \textit{expert} retrieval systems using a ViT-B/16 CLIP model \cite{CLIP} and a ResNet50 DINO model \cite{dino} (see \cref{tab:zero-shot-accuracy-expert-retrieval systems} for a zero-shot evaluation on both the downstream fine-grained datasets as well as on the auxiliary datasets used in $A$). 
Note that the cost to search in the auxiliary dataset $A$ is linear with its size and efficient approximate search methods exist \cite{scann, faiss}, however, when the number of images to search is $<10M$ a brute force solution is still very fast since image embeddings can be stored in GPU memory and the search simply involves a matrix multiplication.

Moreover, for a fair comparison with \cite{adaContrast}, only in the TTA experiments, we follow \cite{adaContrast, SHOT} and add a 256-dimensional bottleneck consisting of a fully-connected layer followed by a BatchNorm layer after the pre-trained backbone, and apply WeightNorm on the classifier. We consider the lower dimensional bottleneck as a projection layer and therefore drop the original projection heads used in MoCo \cite{MoCo} without any performance drop.

\subsection{Memory bank}
In this section we describe the memory bank introduced in \cref{sec:pseudo_labeling_consistency_regularization}. We use a memory bank as in \cite{adaContrast, comatch, MoCo} to allow for a larger set of negative examples without requiring more forward passes (and more GPU memory) and to store previously computed logits that are subsequently used to get ``filtered'' pseudo-labels. 

We maintain a memory queue $M$ of size $m_b$ throughout training. $M$ contains both the last layer features $g$ of samples from $T$ and $A$ and the logits of samples from $T$ according to the current model $f_{A|T}$. 
\begin{equation}\label{eqn:memory_bank}
    M =\Big\{g(t^\prime(x)), f_{A|T}(t^\prime(x)) \Big| x \in T \Big\} \cup \Big\{g(t^\prime(x^\prime)) \Big| x^\prime \in A\Big\}
\end{equation}
where $t^\prime$ is a weak augmentation, $T$ the target dataset and $A$ the auxiliary external dataset. Note that we do not keep track of pseudo-labels (or logits) in $A$ since we do not assume that $A$ contains the same label space of $T$. 
$M$ is initialized with features and probabilities of $m_b$ randomly selected target samples. And, at each mini-batch we update $M$ by enqueue and dequeue similar to \cite{MoCo, adaContrast, comatch}, where the momentum encoder is used with $m=0$.
The memory bank is updated on-the-fly with the current mini-batch and, together with features and logits. We also keep track of the unique image IDs that are used to aggregate logits and avoid using same image as negative samples. Furthermore, since our retrieval system $R$ is fixed throughout training, for each sample in $T$ (indexed by its unique ID) we associate a list of nearest neighbors in $A$ that does not change and we use it to efficiently find the nearest neighbours present in $M$ at each iteration.

\section{Expert retrieval systems zero-shot evaluation}\label{sec:retrieval_accuracy_appendix}

In this section we compare the zero-shot performance of our retrievers both on the fine-grained and the auxiliary datasets used to build $A$. 
First, for each dataset in the list of fine-grained/auxiliary datasets we compute image embeddings using an embedding model (ViT-B/16 or ResNet50 DINO) without using any data augmentation. This process is very fast, though it scales linearly with the dataset size, and its cost is essentially the cost of a single forward pass for each image to be indexed. 
Second, we store all the image embeddings in memory (this typically requires 100 times less memory than storing an image at $224\times244$ resolution) as well as their labels. 
Third, for each test image on a given dataset we compute its k-NNs from the list of embeddings and aggregate the corresponding labels (majority voting).

In \cref{tab:zero-shot-accuracy-expert-retrieval systems}, we compare the retrieval models with expert models that are trained on each dataset independently. Note that the CLIP model is strictly better than DINO in zero shot Top1 accuracy and it also competes with a fully trained model on multiple datasets. As we observed in \cref{tab:CLIP_vs_DINO_vs_random_retrievals} the stronger the retrieval the better the performance since more discriminative negative samples are used in the contrastive objective \Cref{eqn:contrastive_loss}. 

\begin{table}[h!]
\centering
\caption{\textbf{How expert are the retrieval systems?} Retrievals Top1 Accuracy (\%) on the best number of k-NNs. We treat the number of NNs as a hyper-parameter and report the highest accuracy results where $k$ is selected on a small validation dataset ($10\%$ of the training dataset). Furthermore, we report the accuracy of strong experts models ResNet50 (RN50) and ResNet101 (RN101) trained independently on each dataset of the external pool \cite{model_zoo}.}
\begin{tabular}{lccc} \toprule
    {Dataset}                & CLIP  & DINO & Expert \cite{model_zoo} \\ 
    \midrule
        \multicolumn{4}{l}{External data pool} \\
    \midrule
    ImageNet1k               &  72.5 & 73.1 & \textbf{77.5} {\tiny (RN101)} \\
    iNaturalist              &  41.3 & 38.1 & \textbf{75.4} {\tiny (RN101)} \\
    Food101                  &  \textbf{89.1} & 67.1 & 88.0 {\tiny (RN101)} \\
    NWPU-RESISC 45           &  93.0 & 88.3 & \textbf{96.5} {\tiny (RN101)} \\
    Logo 2k                  &  \textbf{83.8} & 35.6 & 78.5 {\tiny (RN101)} \\
    \midrule
        \multicolumn{4}{l}{Fine-grained datasets} \\
    \midrule
    Stanford Cars            & 72.3 & 21   & \textbf{93.4} {\tiny (RN50)} \\
    Stanford Dogs            & 70.9 & 68.4 & \textbf{92.0} {\tiny (RN50)} \\
    CUB200                   & 68 & 67.8   & \textbf{78.3} {\tiny (RN50)} \\
    MIT67                    & \textbf{86} & 71.6   & 78.9 {\tiny (RN50)} \\
    FGVC-Aircrafts           & 45.5 & 36   & \textbf{85.4} {\tiny (RN50)} \\
    \bottomrule
\end{tabular}
\label{tab:zero-shot-accuracy-expert-retrieval systems}
\end{table}

\begin{figure}[!h]
\setcounter{figure}{5}
\vspace{-0.2cm}
    \centering
    \includegraphics[width=\linewidth]{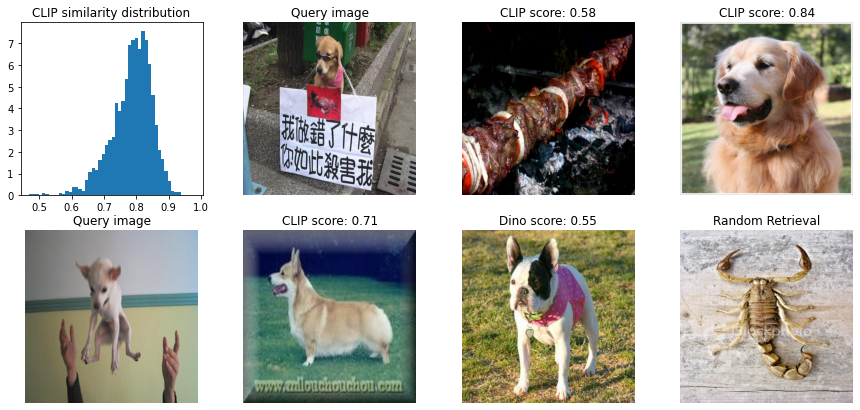}
    \caption{
        \textbf{Visualization of samples retrieved from $A$.} 
        \textit{Top row:} First panel, distribution of CLIP similarity scores on $A$ for the given query image. 
        Third image, low CLIP score, forth image high similarity.
        \textit{Bottom row:} 7-th closest image to the given query according to CLIP and DINO. 
        Their top ranked images are often the same but DINO's ranking gets worse faster. Indeed, on the datasets composing $A$, DINO is a weaker zero-shot retrieval system than CLIP \cref{tab:zero-shot-accuracy-expert-retrieval systems}.
    }
    \label{fig:fpl}
\vspace{-0.5cm}
\end{figure}

\begin{table*}[h!]
\centering
\caption{\textbf{Detailed hyper-parameters configurations.} 
\textit{Ref.\ }refers to the experiment (Table and adaptation method) where the model is mentioned.
\textit{Pre-tr.\ Arch.\ }describes the architecture and the pre-training objective used. 
\textit{Pre-tr.\ data} refers to the pre-training data used to build the pre-trained architecture. 
\textit{Target data} refers to the downstream classification dataset used for evaluation. For TTA it is the same as the pre-training dataset. 
\textit{LR} is the base learning rate (with linear ramp-up and cosine decay and SGD with momentum 0.9). \textit{WD} is weight decay. 
\textit{MixUp} refers to the amount of data augmentation used, where 0. corresponds to no Mixup while 1. is the maximum amunt of data augmentation allowed. 
\textit{Batch Size} is the batch size considered (splitted across multiple GPUs, 8 Tesla V100).}
\begin{tabular}{lcllccccc} 
\toprule
    \midrule
        \multicolumn{9}{c}{Train-Time Adaptation} \\
        Ref & Pre-tr.\ Arch.\  & Pre-tr.\ data & Target data & LR & WD & Mixup\ & Batch Size & Epochs  \\ 
    \midrule
        \cref{tab:train_adaptation_comparison}, Sup.\ FT  \footref{hps_data_splitting} & Sup.\ RN50 \footref{hps_pretraining_method} & IN1k & Stanf.\ Cars  & 0.1 & 0.01  &  0.1  & 1024 & 100 \\ 
        \cref{tab:train_adaptation_comparison}, Sup.\ FT  \footref{hps_data_splitting} & Sup.\ RN50 \footref{hps_pretraining_method} & IN1k & Aircrafts     & 0.1 & 0.01  &  0.1  & 1024 & 100 \\ 
        \cref{tab:train_adaptation_comparison}, Sup.\ FT  \footref{hps_data_splitting} & Sup.\ RN50 \footref{hps_pretraining_method} & IN1k & CUB200        & 0.1 & 0.01  &  0.   & 1024 & 100 \\ 
        \cref{tab:train_adaptation_comparison}, Sup.\ FT  \footref{hps_data_splitting} & Sup.\ RN50 \footref{hps_pretraining_method} & IN1k & MIT-67        & 0.1 & 0.01  &  0.1  & 1024 & 100 \\ 
        \cref{tab:train_adaptation_comparison}, Sup.\ FT  \footref{hps_data_splitting} & Sup.\ RN50 \footref{hps_pretraining_method} & IN1k & Stanf.\ Dogs  & 0.01 & 0.01  &  0.  & 512  & 100 \\
        \cref{tab:train_adaptation_comparison}, \ourname \footref{hps_data_splitting} & Sup.\ RN50 \footref{hps_pretraining_method} & IN1k & Stanf.\ Cars  & 0.1 & 1e-4  &  0.  & 1024 & 100 \\ 
        \cref{tab:train_adaptation_comparison}, \ourname  \footref{hps_data_splitting} & Sup.\ RN50 \footref{hps_pretraining_method} & IN1k & Aircrafts     & 0.1 & 1e-4  &  0.  & 1024 & 100 \\ 
        \cref{tab:train_adaptation_comparison}, \ourname  \footref{hps_data_splitting} & Sup.\ RN50 \footref{hps_pretraining_method} & IN1k & CUB200        & 0.1 & 1e-4  &  0.   & 1024 & 100 \\ 
        \cref{tab:train_adaptation_comparison}, \ourname  \footref{hps_data_splitting} & Sup.\ RN50 \footref{hps_pretraining_method} & IN1k & MIT-67        & 0.1 & 1e-4  &  0.  & 1024 & 100 \\ 
        \cref{tab:train_adaptation_comparison}, \ourname  \footref{hps_data_splitting} & Sup.\ RN50 \footref{hps_pretraining_method} & IN1k & Stanf.\ Dogs  & 0.01 & 1e-4  &  0.  & 512  & 100 \\
    \midrule
        \multicolumn{9}{c}{Test-Time Adaptation} \\
        Ref & Arch.\  & Pre-tr.\ data & Target data  & LR & WD & MixUp\ & Batch Size & Epochs  \\ 
    \midrule
        \cref{tab:test_time_adaptation_for_different_datasetsizes}, \ourname  \footref{hps_data_splitting} & Sup.\ RN50 \footref{hps_pretraining_method_TTA} & DomainNet-126 & DomainNet-126  & 0.1 & 1e-4  &  0.  & 1024  & 30 \\
        \cref{tab:test_time_adaptation_for_different_datasetsizes}, \ourname  \footref{hps_data_splitting} & Sup.\ RN50 \footref{hps_pretraining_method_TTA} & VisDA-C & VisDA-C  & 0.1 & 1e-4  &  0.  & 1024  & 30 \\
    \bottomrule
\end{tabular}
\label{tab:hyper-parameters}
\end{table*}

\section{Tuning details}\label{sec:hps_appendix}
In this section we report the hyper-parameters that we use both for our training-time and test-time adaptation experiments.
In particular, for train-time experiments we used the standard 80/20 train/val splitting, and for test-time experiments we follow \cite{adaContrast}.

\subsection{Train-time model adaptation with retrieval}
In our train-time experiments we consider a labelled dataset $S$ and a large auxiliary dataset of images $A$ (see \cref{sec:experimental_setup}). The goal is to adapt a generic pre-trained model to the downstream labelled task $S$. The performance is evaluated on held out data $T$ that is not used for further adaptation.
This mimics the typical model customization scenario (transfer learning \cite{model_zoo, LQF}) solved with supervised fine-tuning. We pick $S$ to be a labelled fine-grained classification dataset from the ones listed in \cref{sec:experimental_setup}.

In this scenario, we optimize our models\footnote{\label{hps_pretraining_method} For consistency, we keep the same hyper-parameters even when using the self-supervised pre-trained ResNet50 from \cite{dino}.} 
across different datasets\footnote{\label{hps_data_splitting} In case of Sup.\ FT 20\%, to reduce the risk of over-fitting, we decrease the number of epochs to 30 and consider an halved batch size.} 
using SGD with momentum 0.9 and we use linear warm-up cosine annealing learning rate (we use 4 warm-up epochs, start learning rate 1e-5 and minimum learning rate 1e-6), other hyper-parameters are reported in \cref{tab:hyper-parameters}.
We fix $m_b$ to $16k$ samples.

\subsection{Test-time model adaptation with retrieval}
In our test-time experiments we evaluate how well a given pre-trained model can adapt using an unlabelled dataset $T$. In particular, we are given a labelled dataset $S$ that represents the downstream task and has the same label space as $T$ but its covariates are shifted. 
As in previous experiments the auxiliary data pool $A$ is taken as the concatenation of the datasets listed in \cref{sec:experimental_setup}. We evaluate downstream performance with Top1 accuracy on different domains for DomainNet-126 and across different classes for VisDA-C. 

Also, in this case we train our models\footnote{\label{hps_pretraining_method_TTA} In the TTA experiment, for consistency with the literature, we do not use backbones pre-trained with self-supervised objectives \cite{dino}.
}
on differently sized target datasets\footnote{\label{hps_data_splitting_TTA} In case of TTA on 1\% and 10\%, to reduce the risk of over-fitting, we decrease the number of epochs to 30 but do not reduce the batch size.}
use SGD with momentum 0.9 and we use linear warm-up cosine annealing learning rate (we use 4 warm-up epochs, start learning rate 1e-5 and minimum learning rate 1e-6), other hyper-parameters are reported in \cref{tab:hyper-parameters}.
Since both target datasets for TTA are larger than the fine-grained used in our train-time experiments, we increase the size of the memory bank to $64k$ samples. While, when working with 1\% and 10\% of $T$ we use $m_b=16k$.

\section{Datasets details}

\paragraph{Train-time adaptation and auxiliary datasets}
We choose both our fine-grained and the auxiliary datasets such that they cover different domains and are publicly available for download. Detailed data statistics are reported in \cref{tab:suppl_datasets}.

\paragraph{Test-time adaptation datasets}
Following previous literature on test-time domain adaptation \cite{adaContrast, tent, MEMO_test_time_adaptation} we use VisDA-C \cite{VISDA} and DomainNet-126 \cite{domainnet} for evaluating our method on TTA and for comparing against baselines. 
Since DomainNet has noisy labels, we follow \cite{adaContrast} and use a subset of it that only contains 126 classes from 4 domains (Real, Sketch, Clipart and Painting). We therefore evaluate our method on 7 domain shifts constructed from these 4 domains. 
Only for VisDA-C we compare the per-class top-1 accuracy, and then aggregate them by averaging.

\begin{figure}
    \centering
    \includegraphics[width=\linewidth]{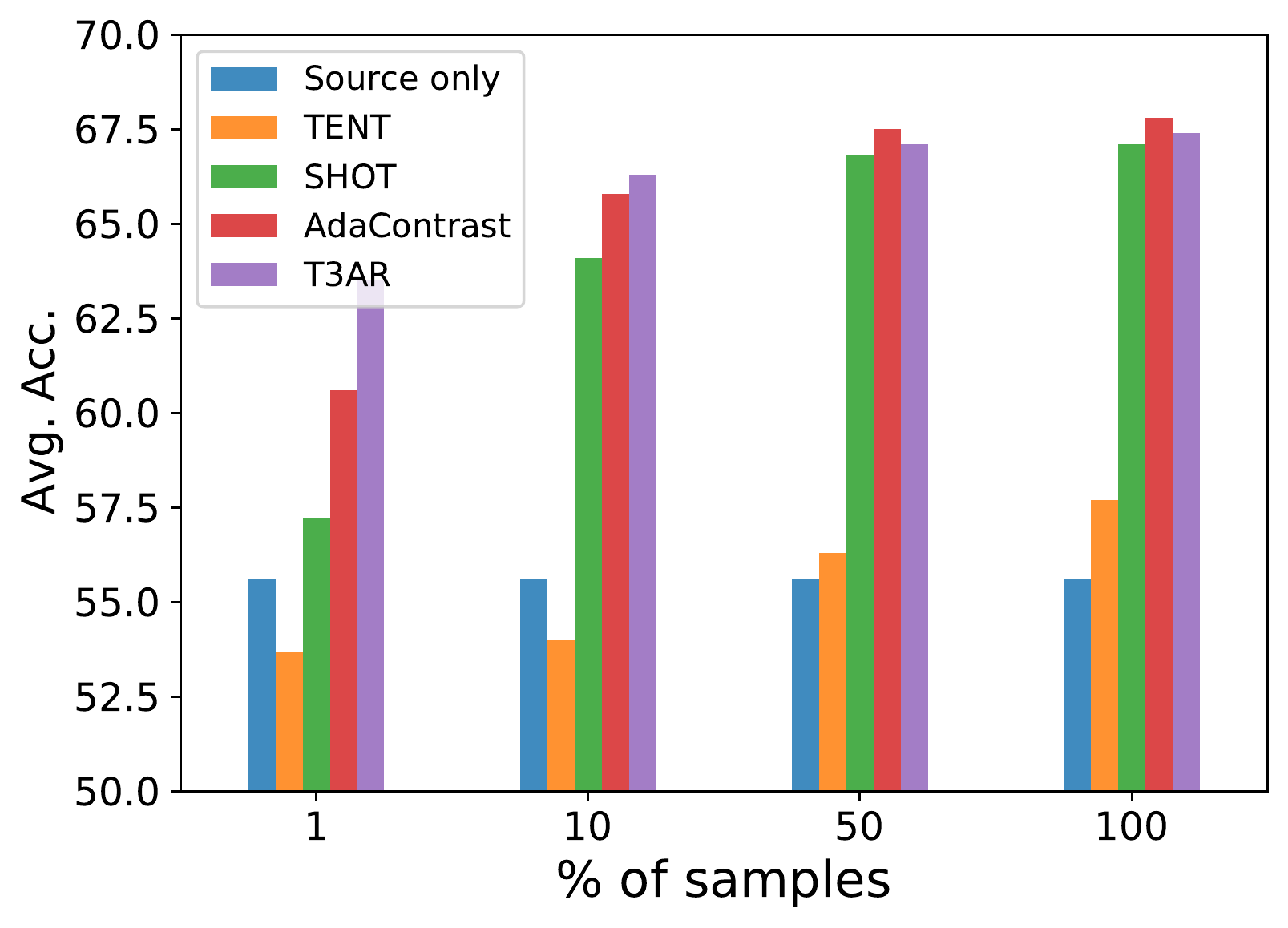}
    \caption{Classification Top1 Accuracy (\%) of test-time adaptation methods on DomainNet-126 as the number of available target data $T$ increases. We show that exploiting retrieved samples helps \ourname especially in the low data regime.}
    \label{fig:T3AR_few_shot}
\end{figure}

\begin{table*}[h!]
\caption{The number of training images, testing images and classes as well as the URL to download the dataset are listed below. The top part contains the auxiliary datasets in $A$, the middle part lists our fine-grained datasets and the bottom part contains test-time adaptation datasets.}
\resizebox{\textwidth}{!}{
\begin{tabular}{lcccl}
\toprule
     {\bf Dataset} & {\bf Training Images} & {\bf Testing Images} & {\bf \# Classes} & {\bf URL} \\
\hline
     NWPU-RESISC45~\cite{nwpu_resisc45} & 25,200 & 6300 & 45 & \footnotesize{\url{https://www.tensorflow.org/datasets/catalog/resisc45}} \\
     Food-101~\cite{food101} & 75,750 & 25,250 & 101 & 
     \footnotesize{\url{https://www.tensorflow.org/datasets/catalog/food101}} \\
     Logo 2k~\cite{logo2k} & 134,907 & 32,233 & 2341 & 
     \footnotesize{\url{https://github.com/msn199959/Logo-2k-plus-Dataset}} \\
     iNaturalist~\cite{iNaturalist_HornASSAPB17} & 265,213 & 3030 & 1010 & 
     \footnotesize{\url{https://github.com/visipedia/inat_comp}} \\
     iMaterialist~\cite{imaterialist} & 965,782 & 9639 & 2019 & 
     \footnotesize{\url{https://github.com/malongtech/imaterialist-product-2019}} \\
     Imagenet~\cite{imagenet_cvpr09} & 1,281,167 & 50,000 & 1000 & 
     \footnotesize{\url{http://image-net.org/download}} \\
\hline
    CUB-200~\cite{WahCUB_200_2011} & 5994 & 5794 & 200 & \footnotesize{\url{http://www.vision.caltech.edu/visipedia/CUB-200-2011.html}} \\ 
    Stanford Cars~\cite{stanfordcars} & 8144 & 8041 & 196 & \footnotesize{\url{https://ai.stanford.edu/~jkrause/cars/car_dataset.html}}\\
    FGVC-Aircrafts~\cite{stanforddogs} & 6667 & 3333 & 100 & \footnotesize{\url{https://www.robots.ox.ac.uk/~vgg/data/fgvc-aircraft/}}\\
    CUB200~\cite{stanforddogs} & 5994 & 5794 & 200 & \footnotesize{\url{https://www.vision.caltech.edu/datasets/cub_200_2011/}}\\
    MIT-67~\cite{stanforddogs} & 5360 & 1340 & 67 & \footnotesize{\url{https://web.mit.edu/torralba/www/indoor.html}}\\
    Stanford Dogs~\cite{stanforddogs} & 12000 & 8580 & 120 & \footnotesize{\url{http://vision.stanford.edu/aditya86/ImageNetDogs}}\\
\hline
    DomainNet-126~\cite{domainnet, adaContrast} & 142334 & - & 126 & \footnotesize{\url{http://ai.bu.edu/M3SDA/}}\\
    VisDA-C~\cite{VISDA} & 152397 & 55388 & 12 & \footnotesize{\url{https://github.com/VisionLearningGroup/taskcv-2017-public}}\\
\bottomrule
\end{tabular}}
\label{tab:suppl_datasets}
\end{table*}

\begin{table*}[t!]
\centering
\caption{Classification accuracy (\%) on 7 domain shifts of DomainNet-126. All methods use ResNet-50 backbone. Bold is the highest. 
Performance of methods from the literature are taken from the cited papers \cite{adaContrast, CAN, MCC, SHOT}. 
In \cref{tab:test_time_adaptation_for_different_datasetsizes} we report the  accuracy as the number of samples available in $T$ decreases.
}
\resizebox{1.6\columnwidth}{!}{
\begin{tabular}{ccccccccc|cc}
    \toprule
    Method & Source-free  & R$\to$C & R$\to$P & P$\to$C & C$\to$S & S$\to$P & R$\to$S & P$\to$R &  Avg. \\
    \hline
    MCC~\cite{MCC} & no & 44.8  &  65.7   & 41.9 & 34.9 & 47.3 & 35.3 & 72.4 &  48.9 \\
    \midrule
    Source only & - & 55.5 & 62.7 & 53.0 & 46.9 & 50.1 & 46.3 & 75.0 &  55.6\\
    TENT~\cite{tent} & yes & 58.5 & 65.7 & 57.9 & 48.5 & 52.4 & 54.0 & 67.0 &  57.7 \\
    SHOT \cite{SHOT} & yes & 67.7 & 68.4 & 66.9 & \textbf{60.1} & \textbf{66.1} & 59.9 & \textbf{80.8} &  67.1 \\
    AdaContrast \cite{adaContrast} & yes & \textbf{70.2} & {69.8} & \textbf{68.6} & 58.0 & 65.9 & \textbf{61.5} & 80.5 &  \textbf{67.8} \\
    \hline
    \ourname (w/o retrievals) & yes & 68.5 & 67.9 & 63.4 & 53.1 & 63.9 & 52.7 & 80.4 & 64.3 \\
    \ourname (Ours) & yes & \textbf{70.2} & \textbf{70.0} & 66.8 & \textbf{60.9} & 64.1 & 59.8 & \textbf{81.0} & 67.5 \\
    \bottomrule
\end{tabular}
}
\label{tab:domainnet_appendix}
\end{table*}

\begin{table*}[t!]
\centering
\caption{Classification accuracy (\%) on VisDA-C train $\to$ val. All methods use ResNet-101 backbone except the on-target rows, which use ResNet-18 as student network. Bold is the highest; underline is the second highest. Performance of methods from the literature are taken from the cited papers \cite{adaContrast, CAN, MCC, SHOT}. In \cref{tab:test_time_adaptation_for_different_datasetsizes} we report the accuracy as the number of samples available in $T$ decreases.
}
\resizebox{1.95\columnwidth}{!}{
\begin{tabular}{cccccccccccccc|cc}
    \toprule
    Method & source-free  & plane & bcycl & bus & car & horse & knife & mcycl & person & plant & sktbrd & train & truck & Avg. \\
    \hline %
    CAN~\cite{CAN} & no  & \underline{97.0} & 87.2 & 82.5 & 74.3 & \textbf{97.8} & \textbf{96.2} & 90.8 & 80.7 & \textbf{96.6} & \textbf{96.3} & 87.5 & \textbf{59.9} & \textbf{87.2} \\
    MCC~\cite{MCC} & no & 88.7 & 80.3 & 80.5 & 71.5 & 90.1 & 93.2 & 85.0 & 71.6 & 89.4 & 73.8 & 85.0 & 36.9 &  78.8 \\
    \hline
    Source only & - & 57.2 & 11.1 & 42.4 & 66.9 & 55.0 & 4.4 & 81.1 & 27.3 & 57.9 & 29.4 & 86.7 & 5.8 &  43.8 \\
    SHOT~\cite{SHOT} & yes  & 95.3 & \underline{87.5} & 78.7 & 55.6 & 94.1 & 94.2 & 81.4 & 80.0 & 91.8 & 90.7 & 86.5 & \underline{59.8} &  83.0 \\
    + On-target & yes & 96.0 & \textbf{89.5} & 84.3 & 67.2 & 95.9 & 94.2 & 91.0 & 81.5 & 93.8 & 89.9 & 89.1 & 58.2 &  85.9 \\
    AdaContrast \cite{adaContrast} & yes  & \underline{97.0} & 84.7 & 84.0 & \underline{77.3} & \underline{96.7} & 93.8 & {91.9} & \underline{84.8} & 94.3 & \underline{93.1} & \textbf{94.1} & 49.7 &  \underline{86.8}  \\
    + On-target & yes & \textbf{97.2} & 87.0 & \textbf{86.7} & \textbf{81.7} & 95.5 & 91.6 & \underline{93.5} & \textbf{86.6} & \underline{95.3} & 90.9 & \underline{92.8} & 47.9 &  \textbf{87.2} \\
    \hline
    \ourname (w/o retrievals) & yes & 90.3 & 83.9 & 72.4 & 73.0 & 93.1 & 88.9 & 82.6 & 82.4 & 90.1 & 87.8 & 90.3 & 40.5 & 81.3 \\
    \ourname (Ours) & yes & 96.8 & \underline{87.5} & \underline{86.2} & 74.8 & \underline{96.7} & 90.5 & \textbf{93.8} & 82.4 & 91.7 & 91.3 & 91.1 & 45.9 & 85.7 \\
    \bottomrule
\end{tabular}
}
\label{tab:visdac_appendix}
\end{table*}

\section{Experiments design}\label{sec:experiments_design_appendix}
In this section we further discuss the main motivations of our experimental study and the main baseline methods we used to evaluate \ourname. 

\paragraph{Fairness of comparison with existing adaptation methods.}
Our experiments are aimed at showing the value of using a new problem formulation for model adaptation that allows retrieval of external information. Hence, rather than aiming at comparing against other algorithms in similar settings, we use existing Train- or Test-Time algorithms to provide strong baselines to quantify what is the value of additional data for downstream adaptation. 
Our main contribution is to show that this setting can significantly improve accuracy, while still being widely applicable (\eg, unlabelled images with a wide domain coverage are readily available from web-scale datasets).

\begin{table}[h!]
    \caption{Comparison of \ourname with a self-supervised model pre-trained on $A$ and then fine-tuned on $S$. We observe that the average accuracy of \ourname outperform the retraining paragon. 
    }
    \centering
    \begin{tabular}{ccccc} \toprule
        {Dataset}                   & Stanf.\ Cars & CUB200 & MIT67 \\ \midrule
        Self-Sup.\@ pre-tr.\ on $A$     &  91.4 & 79.2 & 74.6 \\
        \ourname                    &  \textbf{93.0} & \textbf{80.3} & \textbf{75.9} \\
            \bottomrule
    \end{tabular}
    \label{tab:ablation_pretraining}
\end{table}

\paragraph{Train time experiments}
In the training time experiments we are given a pre-trained model on some pre-training data (pre-trained either with supervision or self-supervision), a labelled dataset $S$ and an unlabelled target dataset $T$. The goal is to adapt the model so that its performance on $T$ is high. This is the standard transfer learning \cite{model_zoo, LQF} setting. We therefore use supervised fine-tuning as strong baselines (with hyper-parameter search \cref{tab:hyper-parameters}). 
However, note that \ourname is allowed to leverage an auxiliary unlabelled dataset $A$ to further improve adaptation.  In general, supervised fine-tuning methods are not designed to exploit side information in $A$. In this case one can resort to semi-supervised techniques to leverage a large set of unlabelled data (\eg  FixMatch \cite{fixmatch} or CoMatch \cite{comatch}). However, these methods cannot be applied in this scenario, since they assume that the labelled dataset $S$ and the unlabelled dataset $A$ are sampled from the same distribution. In our setting, $A$ is more general and does not need to be sampled from the same distribution as $S$ ($A$ can contain many more concepts than the ones present in $S$). 
Therefore, a reasonable baseline is to allow a fine-tuning method to leverage $A$ somehow. The easiest way to do this is by using the whole dataset $A$ to pre-train the model. Since $A$ is, in general, unlabelled, we use a self-supervised objective function (DINO \cite{dino}). Then, this pre-trained model is used as starting checkpoint for fine-tuning on $S$. In this way, the final adapted model contains in its weights both information on the samples from $A$ and samples from $S$ exactly as in the case of \ourname. We call this adaptation strategy \textit{Self-Sup.\@ pre-training} on $A$ and compare it with \ourname in \cref{tab:ablation_pretraining}. Note that \ourname outperforms this paragon. This suggests that, while pre-training a model on as many data as possible is a strong baseline, it is possible to further improve downstream performance by looking back at pre-training data after the downstream ones are available (this observation is aligned with empirical results in \cite{divide_and_contrast}).

\paragraph{Test time experiments}
In the test time experiments we are given a pre-trained model on some labelled dataset $S$ consisting of source data and an unlabelled target dataset $T$. The goal is to adapt the model to $T$ (without using $S$).
This is the standard test-time adaptation setting \cite{adaContrast, tent, SHOT}. We therefore compare \ourname with strong TTA baselines. Furthermore, as typically done in the literature, we add a direct comparison with UDA methods \cite{CAN, MCC}, these allow the method to look back at $S$ once $T$ is obtained. 
The results are reported in  \cref{tab:test_time_adaptation_for_different_datasetsizes}, \cref{tab:domainnet_appendix} and \cref{tab:visdac_appendix}.

\paragraph{Datasets subsampling}
To test the efficacy of external data both in train and test time experiments we tested our method and baselines using both full sized and subsampled downstream datasets. In particular, we subsampled each dataset using stratified sampling.

\section{Detailed results on TTA experiments}\label{sec:detailed_results_appendix}

In \cref{tab:domainnet_appendix} and \cref{tab:visdac_appendix} we report Top1 accuracy on all the domains in DomainNet-126 and on all the classes in VisDA-C. We compare our method with state-of-the-art UDA and TTA methods. 

In \cref{tab:domainnet_appendix} our method outperforms MCC even though it has not access to the source datasets. Our method also compares favourably with TTA baselines, being behind only to AdaContrast on the entire dataset size but being the best when fewer samples are available during adaptation \cref{tab:test_time_adaptation_for_different_datasetsizes}. Our method performs better than others when only 1\% and 10\% of the datasets are allowed for adaptation since it leverages external information from the retrieved samples. In fact, all other methods only rely on synthetic data augmentations to drive the learning process, and therefore, are not fully able to describe the complex target data manifold when data are limited. Interestingly, as more samples are allowed to be used, synthetic data augmentations seems to suffice and the performance of other methods gets increasingly better. We note that our method achieves the best performance on 3 out of 7 domain shifts and it is on par with AdaContrast on one (R$\to$C).

In \cref{tab:visdac_appendix} we compare our method on the VisDA-C adaptation dataset. 
It gets the best accuracies on the \textit{bcycl} class and outperforms AdaContrast (a strong TTA adaptation baseline \cite{adaContrast}) on 4 our of 12 classes. 
Furthermore, \cref{tab:test_time_adaptation_for_different_datasetsizes} we show, once again, that our method compares favourably \wrt the baselines when few samples are allowed for adaptation.

\paragraph{Sensitivity to the number of retrievals}

\begin{figure}
    \centering
    \includegraphics[width=0.8\linewidth]{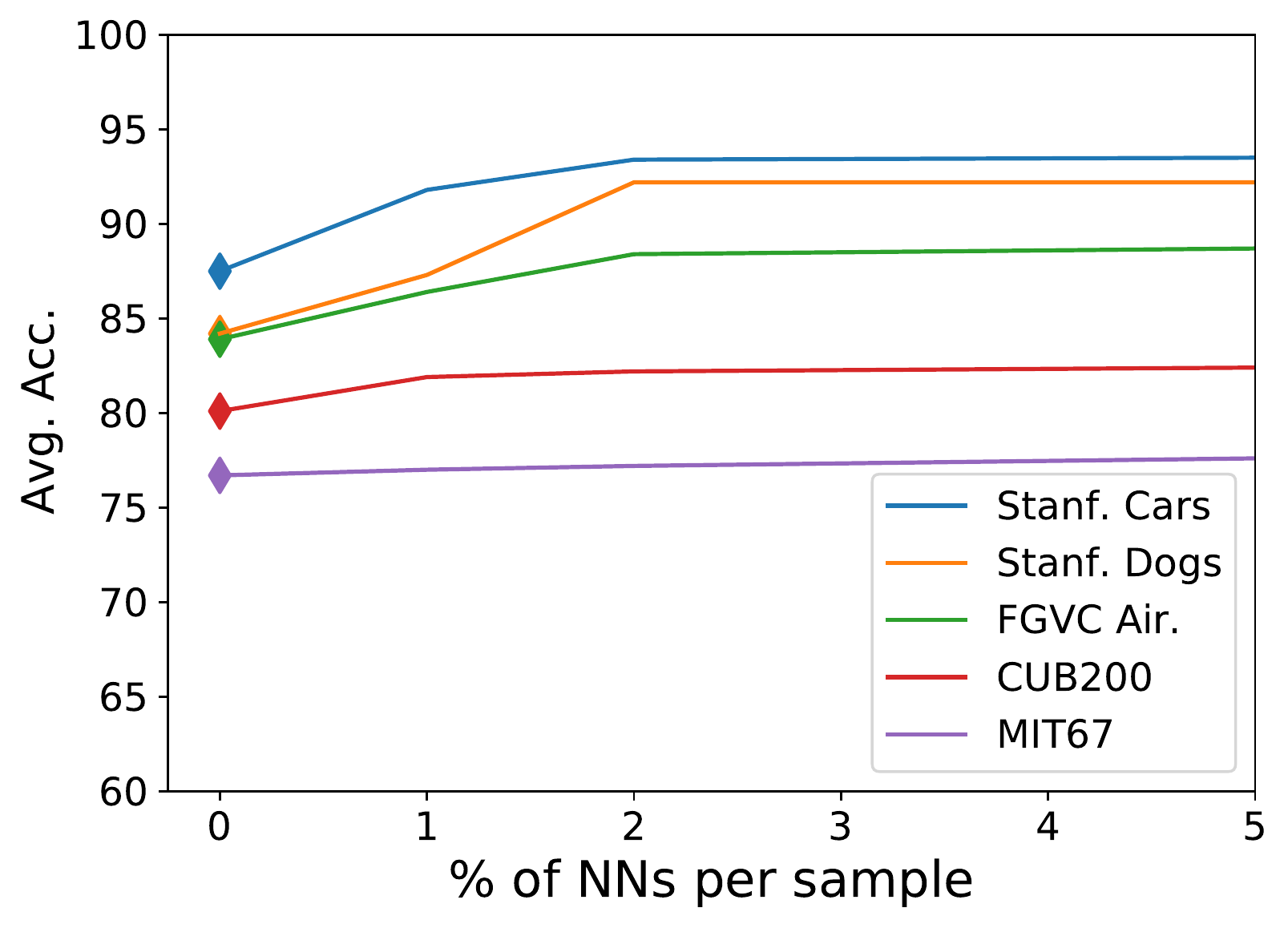}
    \caption{\textbf{Accuracy as a function of the number of the retrievals.} 
    Classification Top1 Accuracy (\%) of \ourname on train time adaptation on fine-grained classifications datasets as a function of the number of retrieved nearest neighbors. We denoted with diamonds the reference performance when random retrievals are used, in this case the number of retrievals is 1. Note that as the number of retrievals increases, as well as the adaptation time, \ourname saturates its performance around 2-5 retrievals across different datasets.}
    \label{fig:T3AR_ablation_num_retrievals}
\end{figure}

In \cref{fig:T3AR_ablation_num_retrievals} we study the sensitivity of \ourname as the number of retrieved nearest neighbors increases. The x-axis represents the number of number of retrievals allowed per sample, with $NNs=1$ we can retrieve as many samples as there are in the target dataset, with $NNs=2$ twice its size, etc.\ We also report the performance of randomly retrieving as many samples as there are in the target dataset (diamond markers at $NNs=0$).
Our results show a diminishing return in performance as the number of NNs increases. 
Since retrieving more samples increases (linearly) adaptation time, our experiments suggest that a good trade-off, that holds across different datasets and allows to discount compute over marginal accuracy improvements, is to retrieve twice as many samples as the target datasets. 

\subsection{Main limitations}
In our ablation studies we have showed that adding samples from $A$ to adapt a downstream model leads to improved downstream performance on various adaptation benchmarks. 
Nonetheless, the user is responsible to bring in relevant data $A$ (as relevant as possible to improve the contrastive loss on negative pairs) and to maintain $A$ as it grows larger and larger. In practice, there is no bound on the size of $A$ and even if similarity based retrievals are very fast, their throughput staturates as more samples are added. We leave to future work how to leverage fast approximate searches \cite{faiss, scann} on large indexed databases and fast database re-indexing. 

\end{document}